\pdfoutput=1

\documentclass[letterpaper, 10 pt, conference]{ieeeconf}

\usepackage[footnotesize]{caption}
\usepackage{subfig}

\usepackage{amsmath}
\usepackage{amssymb}
\usepackage{amsfonts}
\usepackage{graphicx}
\usepackage{psfrag}
\usepackage{tabls}
\usepackage{color}
\usepackage{supertabular}

\usepackage{nccmath}

\usepackage{subfig}

\usepackage{txfonts}

\usepackage{tikz}
\usetikzlibrary{arrows,positioning}
\usetikzlibrary{calc}
\usetikzlibrary{decorations}
\usetikzlibrary{shapes.geometric}

\DeclareMathOperator*{\atantwo}{atan2}

\DeclareMathOperator*{\modulus}{mod}

\begin{document}
\title{\LARGE \bf Recursive Bayesian Initialization of Localization Based on\\ Ranging and Dead Reckoning}

\author{John-Olof Nilsson and Peter H\"{a}ndel\vspace{0mm}\\
{\small Signal Processing Lab, ACCESS Linnaeus Centre}\vspace{-1mm}\\ {\small KTH Royal Institute of Technology}\vspace{-1mm}\\ {\small Osquldas v\"{a}g 10, SE-$100\,44$ Stockholm, Sweden}}

\date{}

\maketitle

\begin{abstract}
  The initialization of the state estimation in a localization scenario based on ranging and dead reckoning is studied. Specifically, we start with a cooperative localization setup and consider the problem of recursively arriving at a uni-modal state estimate with sufficiently low covariance such that covariance based filters can be used to estimate an agent's state subsequently. A number of simplifications/assumptions are made such that the estimation problem can be seen as that of estimating the initial agent state given a deterministic surrounding and dead reckoning. This problem is solved by means of a particle filter and it is described how continual states and covariance estimates are derived from the solution. Finally, simulations are used to illustrate the characteristics of the method and experimental data are briefly presented.
\end{abstract}

\section{Introduction}
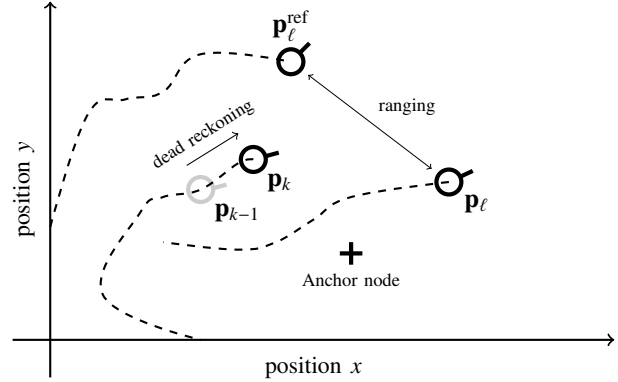
\begin{figure}[t]
\centering
\begin{tikzpicture}[x=1mm,y=1mm]
\draw[->,thick] (-5,0) -- (75,0) node[midway,below=1mm] {\small position $x$};
\draw[->,thick] (0,-5) -- (0,45) {};
\node at (-3.5,20) [rotate=90] {\small position $y$};
\node (agent1) at (32,37) [draw,circle,minimum size=3mm,ultra thick,label={[label distance=-0.5mm]90:$\mathbf{p}^\text{ref}_\ell$}] {};
\draw[ultra thick] ($(agent1) + (45:1.5)$) -- +(45:2);
\node (agent2) at (53,21) [draw,circle,minimum size=3mm,ultra thick,label={[label distance=-1mm]-25:$\mathbf{p}_\ell$}] {};
\draw[ultra thick] ($(agent2) + (25:1.5)$) -- +(25:2);
\node (agent3) at (27,24) [draw,circle,minimum size=3mm,ultra thick,label={[label distance=-1mm]-45:$\mathbf{p}_k$}] {};
\draw[ultra thick] ($(agent3) + (15:1.5)$) -- +(15:2);
\node (agent32) at (20,20) [draw=black!20,circle,minimum size=3mm,ultra thick,label={[label distance=-1mm]-45:$\mathbf{p}_{k-1}$}] {};
\draw[black!20,ultra thick] ($(agent32) + (5:1.5)$) -- +(15:2);
\draw[dashed,thick] plot[smooth,tension=0.7] coordinates {(0,15)  (5,30) (10,32) (15,33) (20,38) (agent1)};
\draw[dashed,thick] plot[smooth,tension=0.7] coordinates {(agent2) (40,19) (35,16) (30,13) (25,12) (15,13)};
\draw[dashed,thick] plot[smooth,tension=0.7] coordinates {(agent3) (25,23.5) (agent32) (15,19) (12,17) (7,6) (20,0)};
\draw[<->,ultra thin,shorten <=1mm, shorten >=1mm] (agent1) -- (agent2) node[midway,right,yshift=2mm] {\scriptsize ranging};
\draw[->] ($(agent32) + (120:3.7mm)$) -- ($(agent3) + (120:4mm)$) node[midway,above,sloped] {\scriptsize dead reckoning};
\draw[-,ultra thick] (40,13) -- (40,10);
\draw[-,ultra thick] (38.5,11.5) -- (41.5,11.5);
\node at (40,8) {\scriptsize Anchor node};
\end{tikzpicture}
\caption{Illustration of the considered localization setup. Multiple agents are localizing cooperatively in three dimensions by means of individual dead reckoning and ranging to other agents or anchor nodes. In this article we suggest a method for initializing the related localization estimation problem.}\label{fig:setup}
\vspace{-5mm}
\end{figure}

Cooperative localization is a highly desired ability in many fields~\cite{Wymeersch2009}\cite{Zhou2008}\cite{Nilsson2013}. At its core is the problem of recursively estimating the involved agents' positions. Commonly, this is done based on dead reckoning and ranging between agents or to anchor nodes/beacons~\cite{Bahr2009}\cite{Nilsson2013}\cite{Zhou2008}. This localization setup is illustrated in Fig.~\ref{fig:setup}. Because of the relative measurements, the errors of the state estimates of different agents may be strongly correlated. Therefore, some joint state estimation is preferable~\cite{Nerurkar2010}\cite{Mazuelas2011}. Due to the resulting high state dimensionality, covariance-based filters (Kalman filters and their derivatives) with relatively low complexity are preferably used. Unfortunately, because of system non-linearities and the periodicity of the orientation, when there are large multi-modal uncertainties in the system, such as during startup or when a new agent enters the localization system, covariance-based filters may give erroneous results. Therefore, some initialization procedure is necessary~\cite{Trawny2010}. Most methods described in the literature use (iterative) least-square solutions~\cite{Zhou2008}\cite{Trawny2007}\cite{Mourikis2006}. However, assuming that recursive Bayesian covariance-based filtering is used after the initialization, employing such techniques would appear somewhat statistically incoherent. Further, the flexibility of varying the cost functions for the suggested methods are low and they do not provide state and covariance estimates throughout the initialization. Therefore, in this article we propose a Bayesian multihypothesis initialization method which recursively arrive at a uni-modal estimate of the agents state, such that a covariance-based filter can be used from there on. The suggested method provide on-line state estimates during the initialization and features the possibility to use an arbitrary likelihood-function for the range measurements, thereby providing the potential to make it statistically robust. Further, the method does not assume any prior or subsequent absolute heading information (e.g. from a magnetometer), but may include it. The methods does not rely on any cooperative feature and therefore it will work equally well for the scenario of ranging relative anchor nodes/beacons only. In general, since the ranging only gives relative measurements, the estimation problem could be turned around and the method used for landmark position initialization~\cite{Blanco2008}. However, this possibility is not further investigated. The basis of the method and the main contribution of this article is the formulation of the initialization problem as that of estimating the starting state rather than that of estimating the current state.
%

\emph{Reproducible research:} A Matlab implementation of the suggested method and code for reproducing the simulations in this article are available at \emph{\color{blue} www.openshoe.org}.

\section{Localization setup}\label{sec:setup}
A number of agents perform dead reckoning. The state $\mathbf{x}_k$ of an agent is its position in three dimensions $\mathbf{p}_k=[x_k,y_k,z_k]$ and heading in the horizontal plane $\theta_k$, i.e. $\mathbf{x}_k=[\mathbf{p}_k,\theta_k]$, where $[\,\cdot\,,\dots]$ is used to denote a column vector. Accordingly, the state space (dead reckoning) model of an agent is \begin{equation}\label{eq:dead_reckoning}
\mathbf{x}_k=\mathbf{x}_{k-1}+\mathbf{R}(\theta_{k-1})\left(\mathbf{u}_k+\mathbf{w}_k\right)
\end{equation}
where $k$ is a time index, $\mathbf{u}_k=[d\mathbf{p}_k,d\theta_k]=[dx_k,dy_k,dz_k,d\theta_\ell]$ is the measured displacement in three dimensions and heading change in the horizontal plane in the agent frame,
\begin{equation*}
\mathbf{R}(\theta_{k})=
\begin{bmatrix}
\cos(\theta_k) & -\sin(\theta_k) & 0 & 0 \\
\sin(\theta_k) & \cos(\theta_k) & 0 & 0 \\
0 & 0 & 1 & 0 \\
0 & 0 & 0 & 1 \\
\end{bmatrix}
\end{equation*}
is the rotation matrix from the agent frame to the navigation coordinate frame, and $\mathbf{w}_k$ is a (by assumption) white error in $\mathbf{u}_k$ with covariance $\mathbf{Q}_k$. 

In addition to the dead reckoning, the agents perform ranging. Given the position of a considered agent $\mathbf{p}_\ell$ and the position of another agent (who is already initialized) or an anchor node $\mathbf{p}^\text{ref}_\ell$ at some time of ranging $\ell$ (a subset of the $k$ index), the range measurements are modeled by some likelihood function
\begin{equation*}
\tilde{r}_\ell\sim\mathcal{V}\left(\,\tilde{r}\,|\,\|\mathbf{p}_\ell-\mathbf{p}^\text{ref}_\ell\|\,\right)
\end{equation*}
where $\sim$ denotes that $\tilde{r}_\ell$ is a sample from the distribution $\mathcal{V}(\cdot)$. 
Often a model with additive noise $\tilde{r}_\ell=\|\mathbf{p}_\ell-\mathbf{p}^\text{ref}_\ell\| + v_\ell$, where $v_k\sim\mathcal{V}(\tilde{r}|0)$ can be seen, but in the suggested initialization method any evaluable likelihood function will suffice. See Section~\ref{subsec:rngud} for the usage of $\mathcal{V}(\cdot)$. This gives a large flexibility when it comes to tweaking the statistical properties of the method to make it robust. \emph{Note that from the initialization method's perspective, there is no difference between another agent and an anchor node.}

\section{Initialization}\label{sec:init}
Based on initial range measurements $\{\tilde{r}_0,\tilde{r}_{1},\dots\}$ and dead reckoning data $\{\mathbf{u}_1,\mathbf{u}_2,\dots\}$ of an agent who has joined the cooperative localization, our desire is to arrive at a uni-modal estimate of $\mathbf{x}_k$ with sufficiently low uncertainty such that the mean $\hat{\mathbf{x}}_k$ and error covariance $\mathbf{P}_k$ estimates together with covariance-based filtering can be used to carry on the localization of the agent relative to the other agents in the system. In addition, we wish to provide continual mean $\hat{\mathbf{x}}_k$ and covariance $\mathbf{P}_k$ estimates throughout the initialization and to determine suitable termination conditions for the initialization. To indicate which range measurements an estimate has been conditioned on, a second subscript $\ell$, as in $\hat{\mathbf{x}}_{k|\ell}$ is added where appropriate.

The uncertainties in the state of an agent will initially be high (infinite in the position domain before any range measurement has taken place) such that some multihypothesis filtering will be necessary during the initialization. To make such a filtering feasible, some simplifications are necessary. Therefore, the following assumptions are made:
\begin{enumerate}
\item Only ranging relative initialized agents or anchor nodes will be considered. Potential range measurements relative to other agents under initialization are discarded.
\item The position errors of initialized agents and anchor nodes are assumed small relative to the initialization uncertainties and are, therefore, ignored.
\item The dead reckoning errors (relative errors) are assumed small over the period of the initialization and are therefore, from the perspective of the initialization procedure, ignored.
\end{enumerate}
Consequently, the initialization will be treated as a local estimation problem in which only the state of the current agent is estimated, treating the world around it and its dead reckoning as deterministic. Note that the assumptions 1-3 are often implicitly made for the initialization methods described in the literature~\cite{Zhou2008}\cite{Trawny2007}\cite{Mourikis2006} so they are not unique to the suggested solution.

Assumptions 1 and 2 serve the purpose of decoupling the initialization estimation problem from the remaining cooperative localization. However, assumption 3 may seem unnecessary at first. The straight-forward solution to the estimation problem is to run a particle filter recursively estimating the agents position and heading given the two first assumptions~\cite{Howard2003}. However, with no initial heading information, the required number of particles will be large and propagating the particles with dead reckoning will require evaluating $\cos(\theta_k)$ and $\sin(\theta_k)$ for each particle and update. The computational cost of this may be prohibitive and unnecessary if assumption 3 holds. Also, if multiple dead reckoning systems are used per agent as in~\cite{Nilsson2013}, running a joint particle filter will not be feasible. Instead, as will be explained, with the 3$^\text{rd}$ assumption, the initial state $\mathbf{x}_0$ may be estimated only requiring the trigonometric functions to be evaluated when a particle is resampled. In other words, the 3$^\text{rd}$ assumption makes it possible to apply the multihypothesis estimation on the static $\mathbf{x}_0$ rather than on $\mathbf{x}_k$ which changes with time.
Estimating $\mathbf{x}_0$ rather than $\mathbf{x}_k$ allows us to use a significantly lower number of hypotheses (also referred to as particles). This is because we can transform the dead reckoning to be relative an estimate $\hat{\mathbf{x}}_0$, and this transformation is invertible. Define the frame transformations
\begin{equation}
\begin{split}
T(\hat{\mathbf{x}}_k,\hat{\mathbf{x}}_0)&=\mathbf{R}^\top(\hat{\theta}_0)\,(\hat{\mathbf{x}}_k-\hat{\mathbf{x}}_0)=\hat{\mathbf{x}}^0_k\\
T^{-1}(\hat{\mathbf{x}}^0_k,\hat{\mathbf{x}}_0)&=\mathbf{R}(\hat{\theta}_0)\,\hat{\mathbf{x}}^0_k+\hat{\mathbf{x}}_0=\hat{\mathbf{x}}_k.
\end{split}\label{eq:frame_trans}
\end{equation}
between the frame relative an estimate $\hat{\mathbf{x}}_0$ and the frame relative of an initial state equal to zero denoted with the superscript $(\cdot)^0$. Here $(\cdot)^\top$ indicates the transpose operation. The corresponding covariances may be transformed accordingly.
With the zero-frame as a basis, the dead reckoning may freely be transform to be relative to any estimate of $\mathbf{x}_0$. Consequently, no intermediate (and potentially poor) $\hat{\mathbf{x}}_{0|\ell}$ estimates will have an irreversible effect on the dead reckoning, and therefore, a low number of particles may be used.
Finally, in the resampling (see Section~\ref{subsec:resampling}) a small ''forgetting factor`` may be added in the initialization, ensuring that small errors in the dead reckoning or the position of other agents or anchor nodes, i.e. assumptions 2-3, will not be a problem.

\subsection{Filtering}

During the initialization (as well as subsequently), the dead reckoning \eqref{eq:dead_reckoning} is used to propagate mean $\hat{\mathbf{x}}_k$ and covariance $\mathbf{P}_k$ estimates of $\mathbf{x}_k$ for all $k$ according to
\begin{align*}\label{eq:dead_reckoning2}
\hat{\mathbf{x}}_k&\!=\hat{\mathbf{x}}_{k-1}+\mathbf{R}(\hat{\theta}_{k-1})\mathbf{u}_k\\
\mathbf{P}_k&\!=\!\mathbf{F}(\hat{\theta}_{k-1},\!dx_k,\!dy_k)\mathbf{P}_{k-1}\mathbf{F}^\top\!(\hat{\theta}_{k-1},\!dx_k,\!dy_k)\!+\!\mathbf{R}(\hat{\theta}_{k-1})\mathbf{Q}_k\mathbf{R}^\top\!(\hat{\theta}_{k-1})
\end{align*}
where the system matrix is
\begin{equation*}
\mathbf{F}(\hat{\theta}_{k-1},dx_k,dy_k)=\!
\begin{bmatrix}
1 & 0 & 0 & \!\!-\sin(\hat{\theta}_{k-1})dx_k-\cos(\hat{\theta}_{k-1})dy_k \\
0 & 1 & 0 & \!\cos(\hat{\theta}_{k-1})dx_k-\sin(\hat{\theta}_{k-1})dy_k \\
0 & 0 & 1 & 0 \\
0 & 0 & 0 & 1 \\
\end{bmatrix}.
\end{equation*}
The dead reckoning is done relative to an initial state estimate $\hat{\mathbf{x}}_0$, and therefore, the estimates $\hat{\mathbf{x}}_k$ and $\mathbf{P}_k$ should be continually updated as the initialization's estimate of $\mathbf{x}_0$ is updated. Therefore, the covariance $\mathbf{P}^0_k$ of $\hat{\mathbf{x}}^0_k$ is also tracked during the initialization by
\begin{gather*}\label{eq:double_cov}
\mathbf{P}^0_k\!=\!\mathbf{F}(\hat{\theta}^0_{k-1}\!,\!dx_k,\!dy_k)\mathbf{P}^0_{k-1}\mathbf{F}^\top\!(\hat{\theta}^0_{k-1}\!,\!dx_k,\!dy_k)\!+\!\mathbf{R}(\hat{\theta}^0_{k-1})\mathbf{Q}_k\mathbf{R}^{\!\top}\!(\hat{\theta}^0_{k-1}).
\end{gather*}


Note that unless agents move, there is no dependence on the heading in the system. Consequently, initialization in the current setup requires motion of the initialized agent. 
However, this will be detected by the terminating conditions of the initialization (see Section~\ref{subsec:term}) and it will not terminate until this is the case.

\subsection{Initial state hypotheses}
Before any range measurement relative to the considered agent is given, the initial position (and heading) prior is assumed to be uniform. However, some initial estimate $\hat{\mathbf{x}}_{0|-1}$ is needed and it may be set identical to zero or some other agent's state.
At the first range measurement, the posterior will be identical with the likelihood function. Since this likelihood function has a simple geometry, it can be deterministically sampled to start the multihypothesis estimation.

The initial sampling is done based on a number of base hypotheses. From the nature of the application, e.g. reasonably flat environment, or some exteroceptive sensor such as a barometer, a set of initial height hypotheses $h^{(i)}$ with weights $w_h^{(i)}$ are assumed given. (Uniform hypotheses of the inclination of the agent relative $\mathbf{p}^\text{ref}_0$ could be used instead.) The first ranging $\tilde{r}_0$ gives a set of true range hypotheses $r^{(j)}$ with weights $\mathcal{V}(r^{(j)}|\tilde{r}_0)=w_r^{(j)}$. Hypotheses of the bearing $\chi^{(n)}$ relative to $\mathbf{p}^\text{ref}_0$ are assumed uniformly distributed (over $[0,2\pi)$). The initial heading hypotheses $\theta^{(m)}$ are uniformly distributed with either uniform weights $w_\theta^{(m)}$ or weights according to some prior or external information source such as a magnetometer or similar. From these base hypotheses and the geometry of the setup, initial state hypotheses/particles are given by
\begin{equation*}
\mathbf{x}^{(i,j,n,m)}_0=\bar{\mathbf{x}}^\text{ref}_{\ell}-\bar{\mathbf{r}}^{(i,j,n)}-\mathbf{r}^{(m)}
\end{equation*}
where
\begin{align*}
\bar{\mathbf{x}}^\text{ref}_{\ell}&=[\mathbf{p}^\text{ref}_{\ell_0},\,0]\\
\bar{\mathbf{r}}^{(i,j,n)}&=[\bar{r}^{(i,j)}\cos(\chi^{(n)}),\,\bar{r}^{(i,j)}\sin(\chi^{(n)}),\,z^\text{ref}_\ell\!-\!\hat{z}^0_\ell\!-\!h^{(j)},\,0]\\
\mathbf{r}^{(m)}&=[\mathbf{H}(\theta^{(m)})\,T(\hat{\mathbf{x}}_{\ell},\hat{\mathbf{x}}_{0|0}),\,-\theta^{(m)}]
\end{align*}
where the horizontal projection of the range hypothesis is $\bar{r}^{(i,j)}=(|(r^{(i)})^2-(z^\text{ref}_\ell-\hat{z}_\ell-h^{(j)})^2|)^{1/2}$ and
\begin{equation}
\mathbf{H}(\theta^{(m)})=
\begin{bmatrix}
\cos(\theta^{(m)}) & -\sin(\theta^{(m)}) & 0 & 0 \\
\sin(\theta^{(m)}) & \cos(\theta^{(m)}) & 0 & 0 \\
0 & 0 & 1 & 0 \\
\end{bmatrix}.
\end{equation}
The corresponding particle weights are $w_{|0}^{(i,j,m)}=w_h^{(i)}w_r^{(j)}w_\theta^{(m)}\,\Sigma^{-1}$ where $\Sigma$ is a normalizing factor such that the weights sum up to 1.
An illustration of the geometry giving the initial hypotheses is shown in Fig.~\ref{fig:init_sampling}. Note that the absolute value in $\bar{r}^{(i,j)}$ is necessary to avoid potential problems for small $r^{(i)}$. Since the origin of the particles/hypothesis in terms of the base hypotheses does not matter, from here on, they and their weights will be indexed by a single index as in $\mathbf{x}^{(i)}_0=[\mathbf{p}^{(i)}_0,\theta^{(i)}_0]$ and $w_{|0}^{(i)}$.
Note that, since the initial bearing and heading hypotheses are static, their corresponding $\cos(\cdot)$ and $\sin(\cdot)$ values can be precalculated. 

\begin{figure}[t]
\centering
\begin{tikzpicture}[x=1mm,y=1mm]
\draw[->,thick] (-5,0) -- (75,0) node[midway,below=1mm] {\small position $x$};
\draw[->,thick] (0,-5) -- (0,48) {};
\node at (-3.5,24) [rotate=90] {\small position $y$};
\begin{scope}[yshift=-1mm]
\node (agent1) at (32,37) [draw,circle,minimum size=3mm,ultra thick,label={[label distance=-0.5mm]90:$\mathbf{p}^\text{ref}_\ell$}] {};
\draw[ultra thick] ($(agent1) + (45:1.5)$) -- +(45:2);
\node (agent2) at (53,21) [draw,circle,minimum size=3mm,ultra thick,label={-90:$\mathbf{p}_\ell$}] {};
\draw[ultra thick] ($(agent2) + (45:1.5)$) -- +(45:2);
\node (start) at (16,13) [draw,circle,minimum size=1mm,inner sep=0mm,fill=black,label={[label distance=-1mm]225:$\mathbf{x}^{(i,j,n,m)}$}] {};
\draw[dashed,thick] plot[smooth,tension=0.7] coordinates {(0,15)  (5,30) (10,32) (15,33) (20,38) (agent1)};
\draw[dashed,thick] plot[smooth,tension=0.7] coordinates {(agent2) (40,19) (35,16) (30,13) (25,12) (start)};
\draw[ultra thin] let \p1=($(agent2)-(agent1)$), \n1 = {veclen(\x1,\y1)},\n2 = {atan2(\x1,\y1)} in (agent1) -- +(10mm,0) ++(7mm,0) arc (0:\n2:7mm) node at ($(agent1)+(\n2*0.5+5:11mm)$) {$\chi^{(n)}$};
\draw[thin,->] let \p1=($(agent2)-(agent1)$), \n1 = {veclen(\x1,\y1)},\n2 = {atan2(\x1,\y1)} in (agent1) -- +(\n2:\n1-5mm);
\draw[dotted,thin,gray] let \p1=($(agent2)-(agent1)$), \n1 = {veclen(\x1,\y1)},\n2 = {atan2(\x1,\y1)} in (agent1) ++(\n2-30:\n1-5mm) arc (\n2-30:\n2+30:\n1-5mm) node[above] {$\bar{r}^{(i,j-1)}$};
\draw[thin,->] let \p1=($(agent2)-(agent1)$), \n1 = {veclen(\x1,\y1)},\n2 = {atan2(\x1,\y1)} in (agent1) -- +(\n2:\n1) node[near start,below=2mm] {$\bar{\mathbf{r}}^{(i,j,m)}$};
\draw[dotted,thin,gray] let \p1=($(agent2)-(agent1)$), \n1 = {veclen(\x1,\y1)},\n2 = {atan2(\x1,\y1)} in (agent1) ++(\n2-40:\n1) arc (\n2-40:\n2+40:\n1) node[above] {$\bar{r}^{(i,j)}$};
\draw[thin,->] let \p1=($(agent2)-(agent1)$), \n1 = {veclen(\x1,\y1)},\n2 = {atan2(\x1,\y1)} in (agent1) -- +(\n2:\n1+5mm);
\draw[dotted,thin,gray] let \p1=($(agent2)-(agent1)$), \n1 = {veclen(\x1,\y1)},\n2 = {atan2(\x1,\y1)} in (agent1) ++(\n2-50:\n1+5mm) arc (\n2-50:\n2+50:\n1+5mm) node[above] {$\bar{r}^{(i,j+1)}$};
\draw[thin,->] let \p1=($(start)-(agent2)$), \n1 = {veclen(\x1,\y1)},\n2 = {atan2(\x1,\y1)} in (agent2) -- +(\n2:\n1) node[pos=0.6,above] {$\mathbf{r}^{(m)}$};
\draw[dashed,ultra thin,gray] let \p1=($(start)-(agent2)$), \n1 = {veclen(\x1,\y1)},\n2 = {atan2(\x1,\y1)} in (agent2) ++(\n2-20:\n1) arc (\n2-20:\n2+15:\n1);
\draw[thin,->] let \p1=($(start)-(agent2)$), \n1 = {veclen(\x1,\y1)},\n2 = {atan2(\x1,\y1)} in (agent2) +(\n2+10:\n1-13) -- +(\n2+10:\n1) node[draw,circle,minimum size=1mm,inner sep=0mm,fill=black,label={[label distance=-1mm]\n2+10:$\mathbf{x}^{(i-1,j,n,m)}$},name=hypim1] {};
\draw[thin,->] let \p1=($(start)-(agent2)$), \n1 = {veclen(\x1,\y1)},\n2 = {atan2(\x1,\y1)} in (agent2) +(\n2-10:\n1-13) -- +(\n2-10:\n1) node[draw,circle,minimum size=1mm,inner sep=0mm,fill=black,label={[label distance=-1mm]\n2-10:$\mathbf{x}^{(i+1,j,n,m)}$},name=hypip1] {};
\draw[dashed,thick] plot[smooth,tension=0.7] coordinates {(26,7) (hypim1)};
\draw[dashed,thick] plot[smooth,tension=0.7] coordinates {(23,17.5) (hypip1)};
\end{scope}
\end{tikzpicture}
\caption{Illustration of the geometry in the horizontal plane of the initial state hypothesis sampling. The bearing, height, and range hypotheses give vectors $\bar{\mathbf{r}}^{(i,j,m)}$ which in turn give the relative positions of the agent relative $\mathbf{p}^\text{ref}_{\ell}$. The heading hypotheses and the dead reckoning give the vectors $\mathbf{r}^{(m)}$ between the agent position and the initial position hypotheses $\mathbf{x}^{(i)}$. }\label{fig:init_sampling}
\vspace{-5mm}
\end{figure}

%
%

\subsection{Ranging updates}\label{subsec:rngud}
To condition a particle $\mathbf{x}^{(i)}_0$ (and subsequently $\hat{\mathbf{x}}_{\ell|\ell-1}$ and $\mathbf{P}_{\ell|\ell-1}$) with respect to a range measurement $\tilde{r}_\ell$ where $\ell>0$, it is reweighted with the corresponding likelihood of observing the range measurement. The predicted range according to hypothesis $\mathbf{x}^{(i)}_0$ is
\begin{equation}\label{eq:pred_range}
\hat{r}^{(i)}_\ell=\left\|\mathbf{H}(\theta_0^{(i)})\,T(\hat{\mathbf{x}}_{\ell|\ell-1},\hat{\mathbf{x}}_{0|\ell-1})+\mathbf{p}^{(i)}_0-\mathbf{p}^\text{ref}_\ell\right\|.
\end{equation}
Accordingly, the hypothesis weight conditioned on a range measurement $\tilde{r}_\ell$ is
\begin{equation}\label{eq:reweighting}
w^{(i)}_{|\ell}= w^{(i)}_{|\ell-1}\cdot\mathcal{V}(\tilde{r}_\ell|\hat{r}^{(i)}_\ell)\cdot\Sigma^{-1}
\end{equation}
where again $\Sigma$ is a normalizing factor such that the conditioned weights sum up to 1. Suitably, the likelihood function is taken to be Cauchy-distributed
\begin{equation*}
\mathcal{V}(\,\tilde{r}|\hat{r}_\ell^{(i)}\,) = \frac{\sigma}{\pi}\left(\frac{1}{(\tilde{r}-\hat{r}_\ell^{(i)})^2+\sigma^2}\right)
\end{equation*}
where $\sigma$ is the scale parameter of the distribution. This heavy tailed distribution will make the initialization robust to measurement outliers and is inexpensive to evaluate.

With the conditioned weights, the conditional mean position may be calculated by the weighted sample mean
\begin{equation}\label{eq:mean_pos}
\hat{\mathbf{p}}_{0|\ell}=\medop\sum_i w^{(i)}_{|\ell}\mathbf{p}^{(i)}_{0}.\\
\end{equation}
Unfortunately, due to the periodicity of the heading, the conditional mean cannot be used for the heading. However, since the quality of the estimate is only crucial when the initialization terminates, assumably providing a low variance estimate, the simple vector sum algorithm with weighted components may be used,
\begin{equation}\label{eq:vector_sum}
\hat{\theta}_{0|\ell}=\atantwo\left(\medop\sum_i w^{(i)}_{|\ell}\sin(\theta^{(i)}),\medop\sum_i w^{(i)}_{|\ell}\cos(\theta^{(i)})\right).
\end{equation}
For properties of the vector sum algorithm and more refined methods, see \cite{Ohlson2011}\cite{Kristensen2013}. Together~\eqref{eq:mean_pos} and \eqref{eq:vector_sum} give $\hat{\mathbf{x}}_{0|\ell}=[\hat{\mathbf{p}}_{0|\ell},\hat{\theta}_{0|\ell}]$.

Similar to the conditional mean, the conditional covariance may be calculated by the weighted sample covariance. However, again, care has to be taken to handle the periodicity of the heading. Define the sample deviation from the mean by
\begin{equation*}
\mathbf{e}^{(i)}_{0|\ell}=(\hat{\mathbf{p}}_{0|\ell}-\mathbf{p}^{(i)}_0,\modulus(\hat{\theta}_{0|\ell}-\theta^{(i)}+\pi,2\pi)-\pi)
\end{equation*}
where $\modulus(\cdot,a)$ is the modulus-$a$ division with the sign equal to the divisor. Then the sample covariance is
\begin{equation}
\mathbf{P}_{0|\ell}=\!\medop\sum_i w^{(i)}_{|\ell}\mathbf{e}^{(i)}_{0|\ell}(\mathbf{e}^{(i)}_{0|\ell})^\top.\label{eq:conditional_cov}
\end{equation}

With the conditional mean $\hat{\mathbf{x}}_{0|\ell}$, the current state estimates relative $\hat{\mathbf{x}}_{0|\ell}$ is given by
\begin{equation}
\hat{\mathbf{x}}_{\ell|\ell} = T^{-1}(T(\hat{\mathbf{x}}_{\ell|\ell-1},\hat{\mathbf{x}}_{0|\ell-1}),\hat{\mathbf{x}}_{0|\ell}).
\end{equation}
The corresponding covariance is 
\begin{equation}
\begin{split}
\mathbf{P}_{\ell|\ell}&=\mathbf{F}(\hat{\theta}_{0|\ell},\hat{x}^0_{\ell},\hat{y}^0_{\ell})\mathbf{P}^0_{\ell}\mathbf{F}^\top(\hat{\theta}_{0|\ell},\hat{x}^0_{\ell},\hat{y}^0_{\ell})+\mathbf{P}_{0|\ell},
\end{split}\label{eq:covs}
\end{equation}
i.e. when the covariance in the navigation frame is evaluated, the complete dead reckoning is treated as one step.


\subsection{Initialization termination}\label{subsec:term}
Once a uni-modal distribution of $\mathbf{x}_0^{(i)}$ with sufficiently small covariance has been attained as a result of ranging updates and resampling, the initialization should be terminated and the states of the agent estimated jointly with the remainder of initialized agents in the system. The uni-modality comes together with a small covariance for any reasonable choice of ranging likelihood function.
The small covariance requirement may be assessed by the size of the diagonal entries of $\mathbf{P}_{0|\ell}$.
Consequently, the terminating conditions for the initialization is
\begin{equation*}
\text{diag}(\mathbf{P}_{0|\ell})<\boldsymbol{\gamma}_\text{cov}
\end{equation*}
where $\boldsymbol{\gamma}_\text{cov}$ is the bounds on the allowable variances 
and the 'less-than' relations are applied to each component of the vectors. If this holds, the initialization may be terminated and subsequent range measurement handled by any covariance-based filter of choice.



\subsection{Resampling}\label{subsec:resampling}
After a few ranging updates, most initial particles will have a negligible weight. Unless an excessive number of particles are to be used, this sample impoverishment has to be mitigated by resampling. Ideally we would like to sample new particles from the posterior distribution. Unfortunately, the posterior distribution is only available as a particle cloud. 
However, since a static quantity $\mathbf{x}_0$ is estimated, a simple Gaussian proposal function suffice, and we employ the following resampling strategy. If the weight of a particle is below some threshold
\begin{equation*}
w_{|\ell}^{(i)}<\gamma\frac{1}{N}
\end{equation*}
where $N$ is the number of particles, draw a new $i$th sample
\begin{equation}\label{eq:resampling}
\mathbf{x}^{(i)}_0\sim\mathcal{N}(\hat{\mathbf{x}}_{0|\ell},\alpha^2\,\mathbf{P}_{0|\ell})
\end{equation}
from the Gaussian distribution $\mathcal{N}(\hat{\mathbf{x}}_{0|\ell},\alpha^2\,\mathbf{P}_{0|\ell})$ and make the assignment $w_{|\ell}^{(i)}=1/N$. The threshold $\gamma$ will be a system parameter which will determine the trade-off between particle diversity and particle impoverishment. Most naturally we would use $\alpha=1$. However, if $\alpha>1$ this will add a forgetting factor to the initialization. This can be used to make the initialization robust to errors in the dead reckoning during the initialization at the cost of a potentially prolonged initialization.
%
The resampling \eqref{eq:resampling} is implemented by drawing a sample $\mathbf{n}$ from a standard Gaussian distribution and making the assignment
\begin{equation}
\mathbf{x}_0^{(i)}=\hat{\mathbf{x}}_{0|\ell}+\alpha\,\mathbf{L}_{0|\ell}\cdot\mathbf{n}\label{eq:resampling2}
\end{equation}
where $\mathbf{P}_{0|\ell}=\mathbf{L}_{0|\ell}\mathbf{L}^\top_{0|\ell}$ is the Cholesky factorization of $\mathbf{P}_{0|\ell}$.


\subsection{Computational cost}
Since there is no dynamic in the estimated initial state $\mathbf{x}_0$, the computational cost associated with the initialization only comes from the ranging updates and the resampling and there is no dependence on the number of agents or anchor nodes (apart from that each agent need to be initialized). For each range update and particle, the predicted range $\hat{r}^{(i)}_\ell$ needs to be calculated by~\eqref{eq:pred_range}. The cost is 7 multiplications (mul) and 8 additions and subtractions (add) and a square-root operation. The cost of the reweighting~\eqref{eq:reweighting} (with a Cauchy distributed likelihood function) with a separate renormalization step is approximately 2 mul, 3 add, and one division (div) operation per particle. Calculating the mean position~\eqref{eq:mean_pos} and mean heading~\eqref{eq:vector_sum} will cost 5 mul and 5 add per particle. Calculating the sample deviation $\mathbf{e}^{(i)}_{0|\ell}$ costs 6 add plus a modulus division. In turn, calculating the sample covariance requires 14 mul and 10 add per sample. The computational cost for calculating the transformations~\eqref{eq:frame_trans} and the covariances~\eqref{eq:covs} will be marginal. 

Assuming that samples are drawn from a pre-generated list of samples from a standard Gaussian distribution and thereby do not carry any cost, the computational cost of the resampling \eqref{eq:resampling2} is 40 multiplications and 40 additions per resampled particle. In addition, $\cos(\theta^{(i)})$ and $\sin(\theta^{(i)})$ need to be evaluated and stored for later use for each resampled particle. The Cholesky factorization of the $4\times4$ matrix $\mathbf{P}_{0|\ell}$ only needs to be evaluated once, making the cost marginal. The total number and type of required operations \emph{per particle} is summarized in Table~\ref{tab:cc}.

\begin{table}\caption{Total number of operations required \emph{per particle} for performing the ranging update and resampling in the initialization procedure.}\label{tab:cc}
{\centering \normalsize
\begin{tabular}{r|c|c|c|c}
               & add & mul & div & other \\
\hline
ranging update & 32      & 32  & 1   & $\sqrt{\cdot}$, $\modulus(\cdot,2\pi)$\\
\hline
resampling     & 40      & 40  & -   & $\cos(\cdot)$, $\sin(\cdot)$\\
\end{tabular}}
\end{table}

%

\section{Simulations}\label{sec:sim}
\begin{figure}[t]
\centering
\tikzstyle{agent}=[draw,circle,minimum size=2.5mm,inner sep=0mm,thick,black!30]
\tikzstyle{agent2}=[thick,black!30]
\tikzstyle{ranging}=[<->,ultra thin,shorten <=1mm, shorten >=1mm]
\begin{tikzpicture}[x=0.5mm,y=0.5mm]
\draw[->,thick] (-5,0) -- (160,0) node[midway,below=0mm] {\footnotesize position $x$ [m]};
\draw[->,thick] (0,-5) -- (0,65) {};
\node at (-3mm,27) [rotate=90] {\footnotesize position $y$ [m]};
\draw[-,thick] (150,2) -- (150,0) node[at end,below=0mm] {\footnotesize 150};
\draw[-,thick] (2,50) -- (0,50) node[at end,left=0mm] {\footnotesize 50};
\node (agent10) at (18,33) [agent,label={90:$\mathbf{p}^\text{ref}_0$}] {};
\draw[agent2] ($(agent10) + (-10:1.25mm)$) -- +(-10:1.7mm);
\node (agent11) at (38,37) [agent,label={90:$\mathbf{p}^\text{ref}_1$}] {};
\draw[agent2] ($(agent11) + (45:1.25mm)$) -- +(45:1.7mm);
\node (agent12) at (55,46) [agent,label={90:$\mathbf{p}^\text{ref}_2$}] {};
\draw[agent2] ($(agent12) + (5:1.25mm)$) -- +(5:1.7mm);
\node (agent13) at (70,43) [agent,label={90:$\mathbf{p}^\text{ref}_3$}] {};
\draw[agent2] ($(agent13) + (-25:1.25mm)$) -- +(-25:1.7mm);
\node (agent14) at (89,34) [agent,label={90:$\mathbf{p}^\text{ref}_4$}] {};
\draw[agent2] ($(agent14) + (-5:1.25mm)$) -- +(-5:1.7mm);
\node (agent15) at (105,28) [agent,label={90:$\mathbf{p}^\text{ref}_5$}] {};
\draw[agent2] ($(agent15) + (-15:1.25mm)$) -- +(-15:1.7mm);
\node (agent16) at (134,23) [agent,label={-90:$\mathbf{p}^\text{ref}_6$}] {};
\draw[agent2] ($(agent16) + (5:1.25mm)$) -- +(5:1.7mm);
\node (agent17) at (150,33) [agent,black,label={-90:$\mathbf{p}^\text{ref}_7$}] {};
\draw[agent2,black] ($(agent17) + (45:1.25mm)$) -- +(45:1.7mm);
\draw[dashed,thick] plot[smooth,tension=0.7] coordinates {(0,15) (6,30) (12,32) (agent10) (23,38) (agent11) (42,40) (50,44) (agent12) (59,48) (67,47) (agent13) (75,40) (83,36) (agent14) (91,33) (100,32) (agent15) (108,24) (116,22) (agent16) (141,26) (146,30) (agent17)};
\node (agent20) at (41,16) [agent,label={-90:$\mathbf{p}_0$}] {};
\draw[agent2] ($(agent20) + (5:1.25mm)$) -- +(5:1.7mm);
\node (agent21) at (62,21) [agent,label={-90:$\mathbf{p}_1$}] {};
\draw[agent2] ($(agent21) + (45:1.25mm)$) -- +(45:1.7mm);
\node (agent22) at (80,13) [agent,label={-90:$\mathbf{p}_2$}] {};
\draw[agent2] ($(agent22) + (-15:1.25mm)$) -- +(-15:1.7mm);
\node (agent23) at (96,10) [agent,label={[label distance=-1mm]-45:$\mathbf{p}_3$}] {};
\draw[agent2] ($(agent23) + (0:1.25mm)$) -- +(0:1.7mm);
\node (agent24) at (114,16) [agent,label={[label distance=-1mm]-45:$\mathbf{p}_4$}] {};
\draw[agent2] ($(agent24) + (50:1.25mm)$) -- +(50:1.7mm);
\node (agent25) at (129,37) [agent,label={[label distance=-1mm]120:$\mathbf{p}_5$}] {};
\draw[agent2] ($(agent25) + (45:1.25mm)$) -- +(45:1.7mm);
\node (agent26) at (137,46) [agent,label={90:$\mathbf{p}_6$}] {};
\draw[agent2] ($(agent26) + (55:1.25mm)$) -- +(55:1.7mm);
\node (agent27) at (150,53) [agent,black,label={90:$\mathbf{p}_7$}] {};
\draw[agent2,black] ($(agent27) + (45:1.25mm)$) -- +(45:1.7mm);
\draw[dashed,thick] plot[smooth,tension=0.7] coordinates {(16,13) (29,12) (35,13) (agent20) (47,19) (agent21) (64,20) (79,14) (agent22) (87,8) (95,9) (agent23) (105,9) (111,15) (agent24) (123,23) (127,35) (agent25) (131,40) (135,45) (agent26) (141,50) (146,52) (agent27)};
\draw[ranging] (agent10) -- (agent20) node[midway,right,yshift=2mm] {$\tilde{r}_0$};
\draw[ranging] (agent11) -- (agent21) node[midway,right,yshift=1mm] {$\tilde{r}_1$};
\draw[ranging] (agent12) -- (agent22) node[midway,right,yshift=0.5mm] {$\tilde{r}_2$};
\draw[ranging] (agent13) -- (agent23) node[midway,right,yshift=0mm] {$\tilde{r}_3$};
\draw[ranging] (agent14) -- (agent24) node[midway,below,yshift=0mm] {$\tilde{r}_4$};
\draw[ranging] (agent15) -- (agent25) node[midway,above,yshift=0mm] {$\tilde{r}_5$};
\draw[ranging] (agent16) -- (agent26) node[midway,right,yshift=0mm] {$\tilde{r}_6$};
\draw[ranging] (agent17) -- (agent27) node[midway,right,yshift=0mm] {$\tilde{r}_7$};
\end{tikzpicture}
\caption{Illustration of the simulated trajectories and ranging measurements. The initially lower agent is initialized relative to the upper agent.}\label{fig:sim_trj}
\vspace{-5mm}
\end{figure}
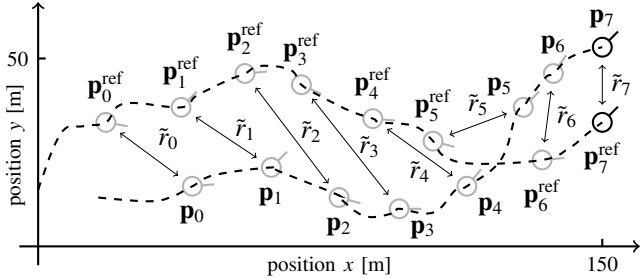

\begin{figure}[t]%
\centering
\resizebox{\linewidth}{!}{\includegraphics{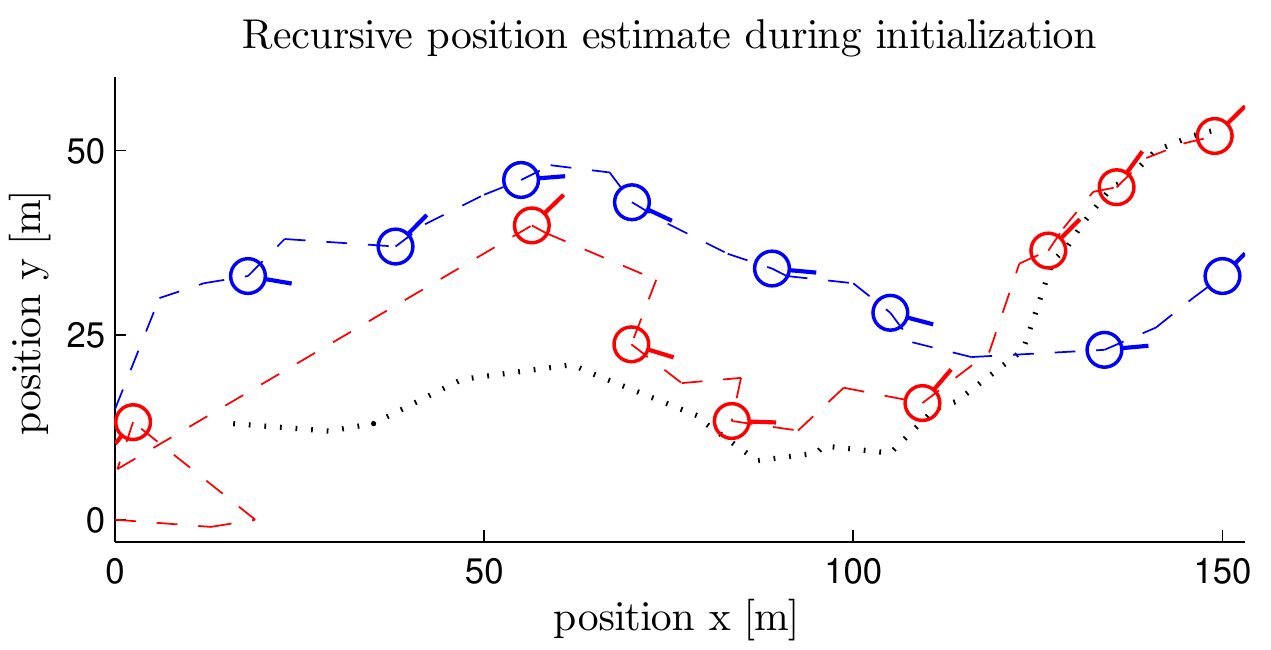}}
\caption{Plot of the recursively estimated agent states. The black dotted line indicate the true trajectory. Initially the state estimate of the lower (red) agent being initialized make large jumps. However, already after a 3 range updates the state estimate is reasonable and after an additional 3 updates, the agreement with the true trajectory is good.}\label{fig:pos_est}
\end{figure}

\begin{figure}[t]%
\centering
\resizebox{0.47\linewidth}{!}{\includegraphics{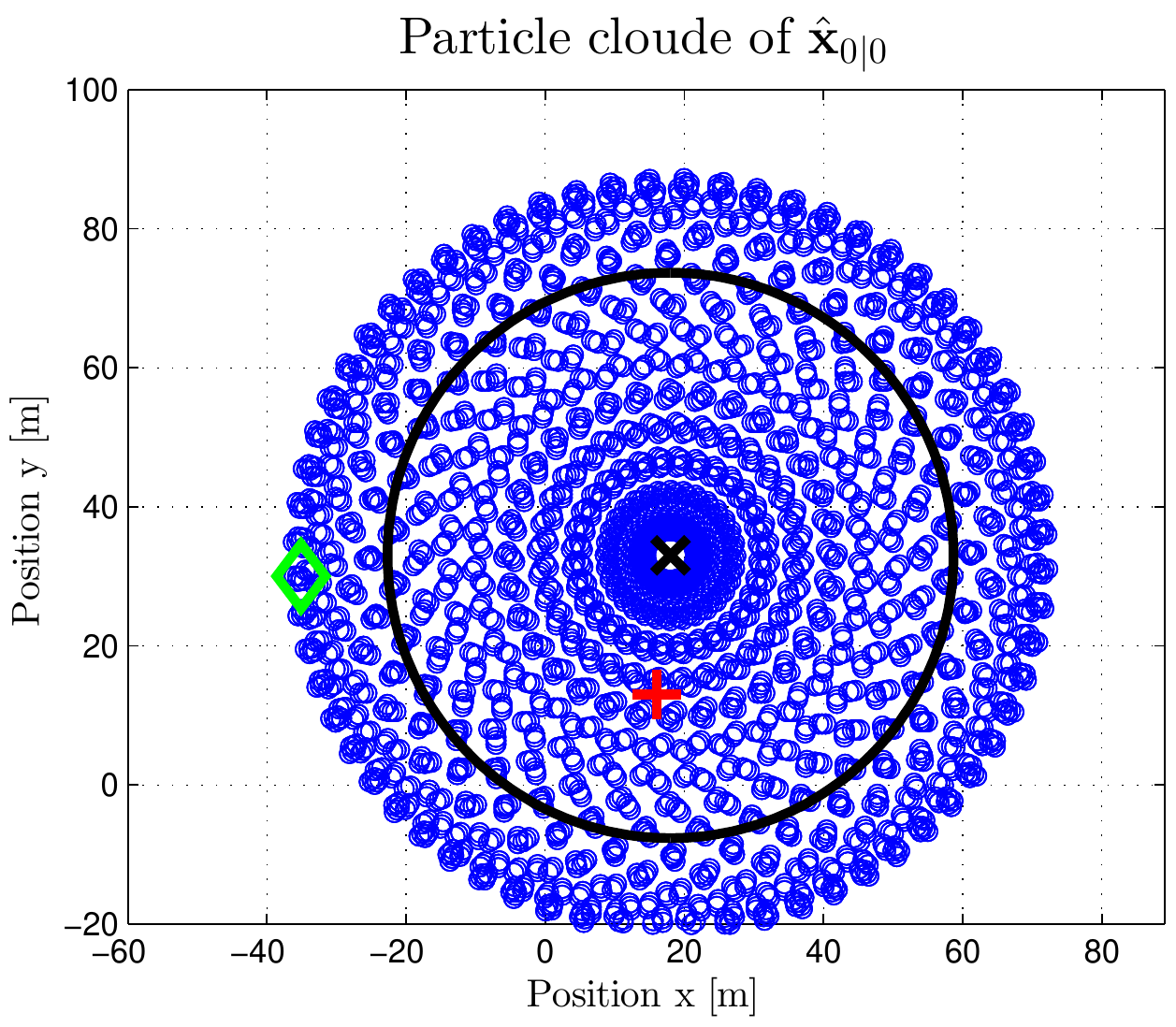}}
\resizebox{0.47\linewidth}{!}{\includegraphics{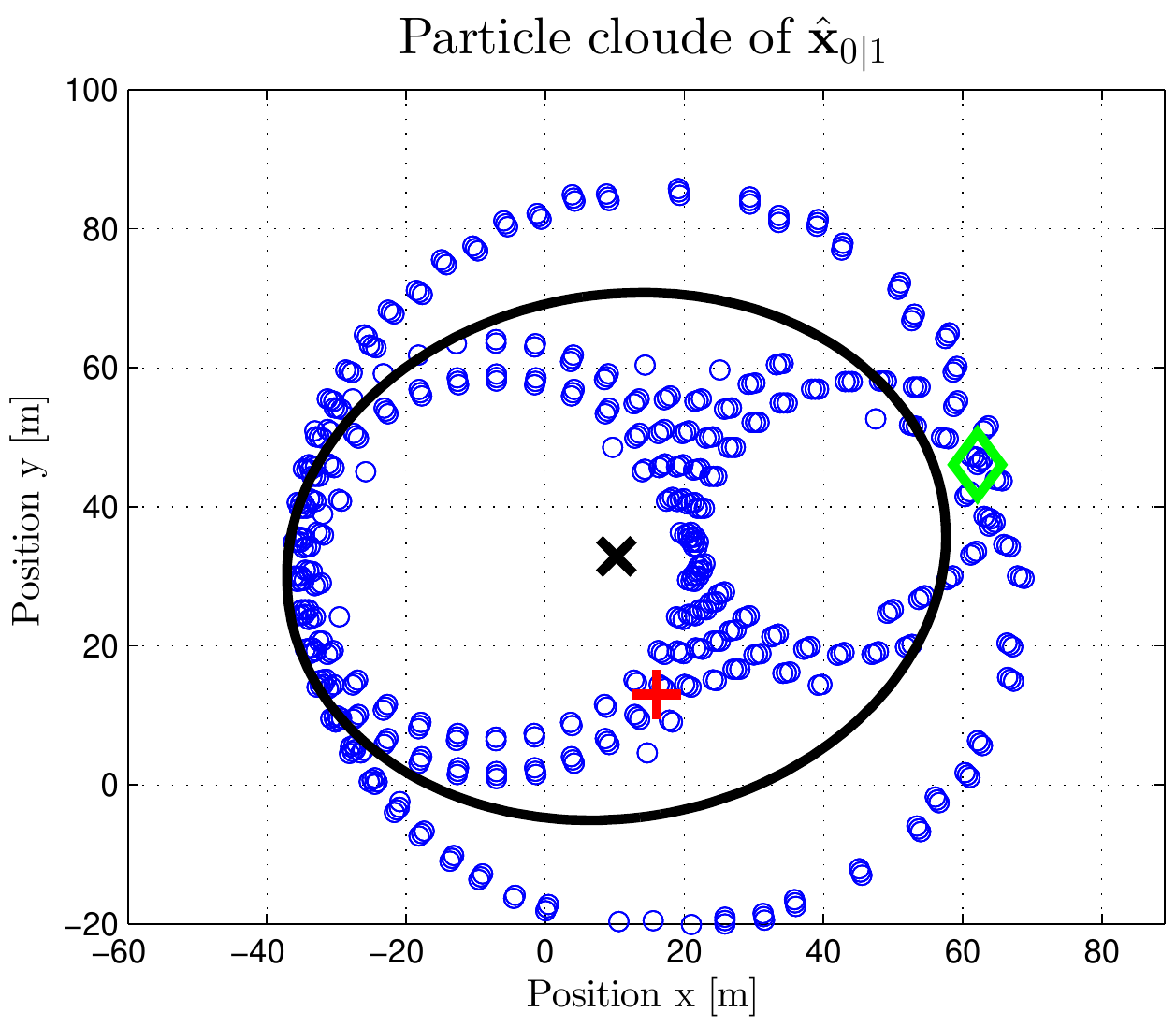}}

\hspace{1.5mm}
\resizebox{0.47\linewidth}{!}{\includegraphics{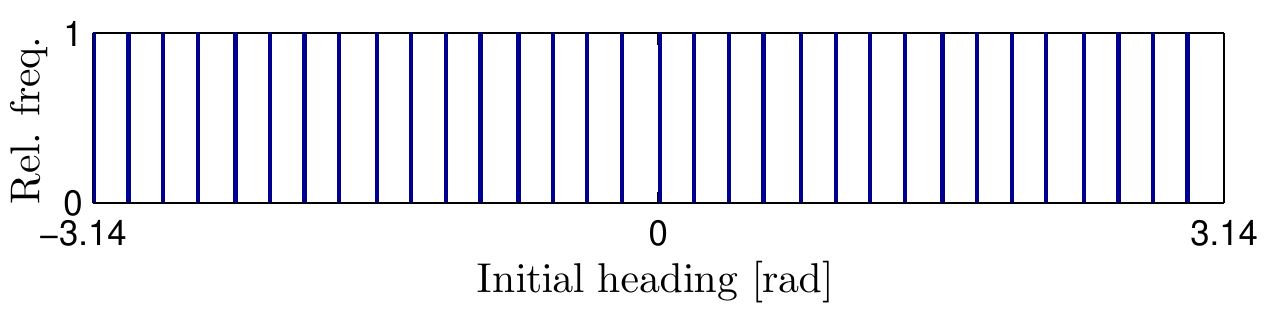}}
\resizebox{0.47\linewidth}{!}{\includegraphics{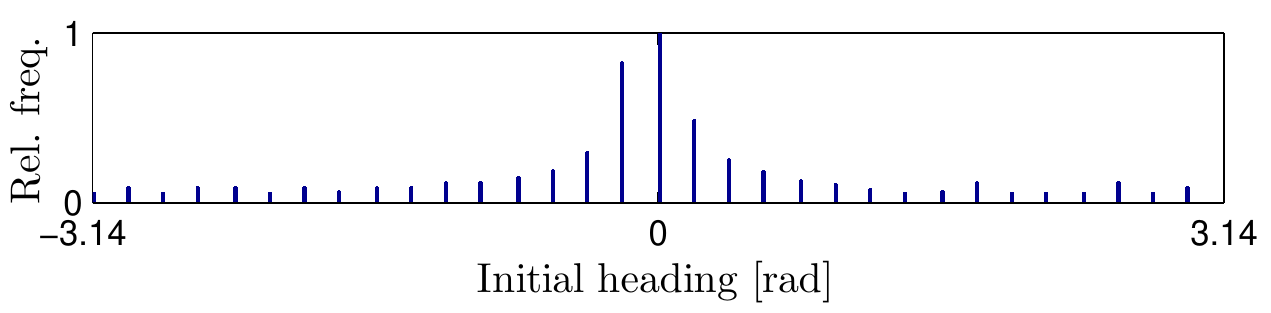}}

\vspace{4mm}
\resizebox{0.47\linewidth}{!}{\includegraphics{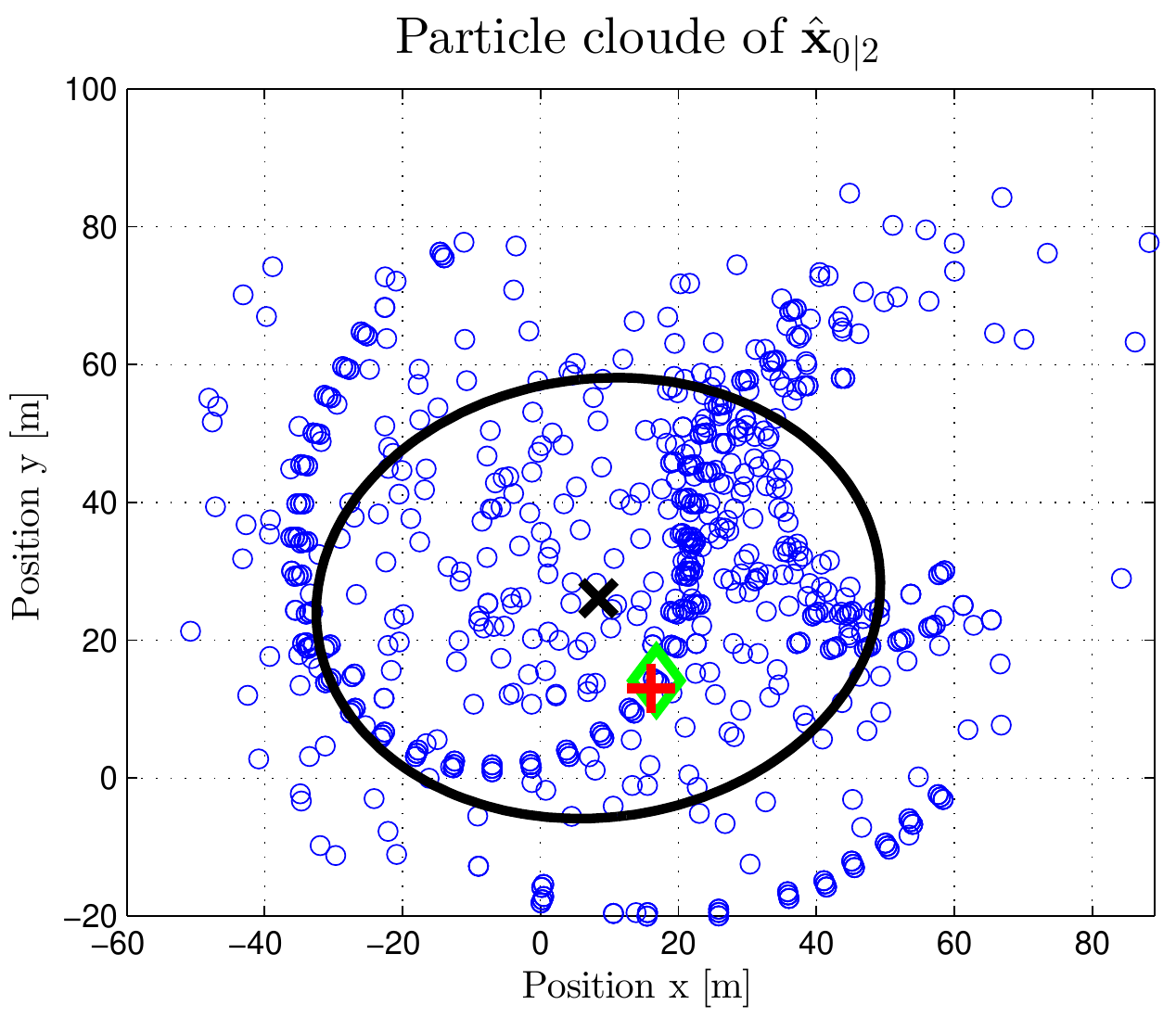}}
\resizebox{0.47\linewidth}{!}{\includegraphics{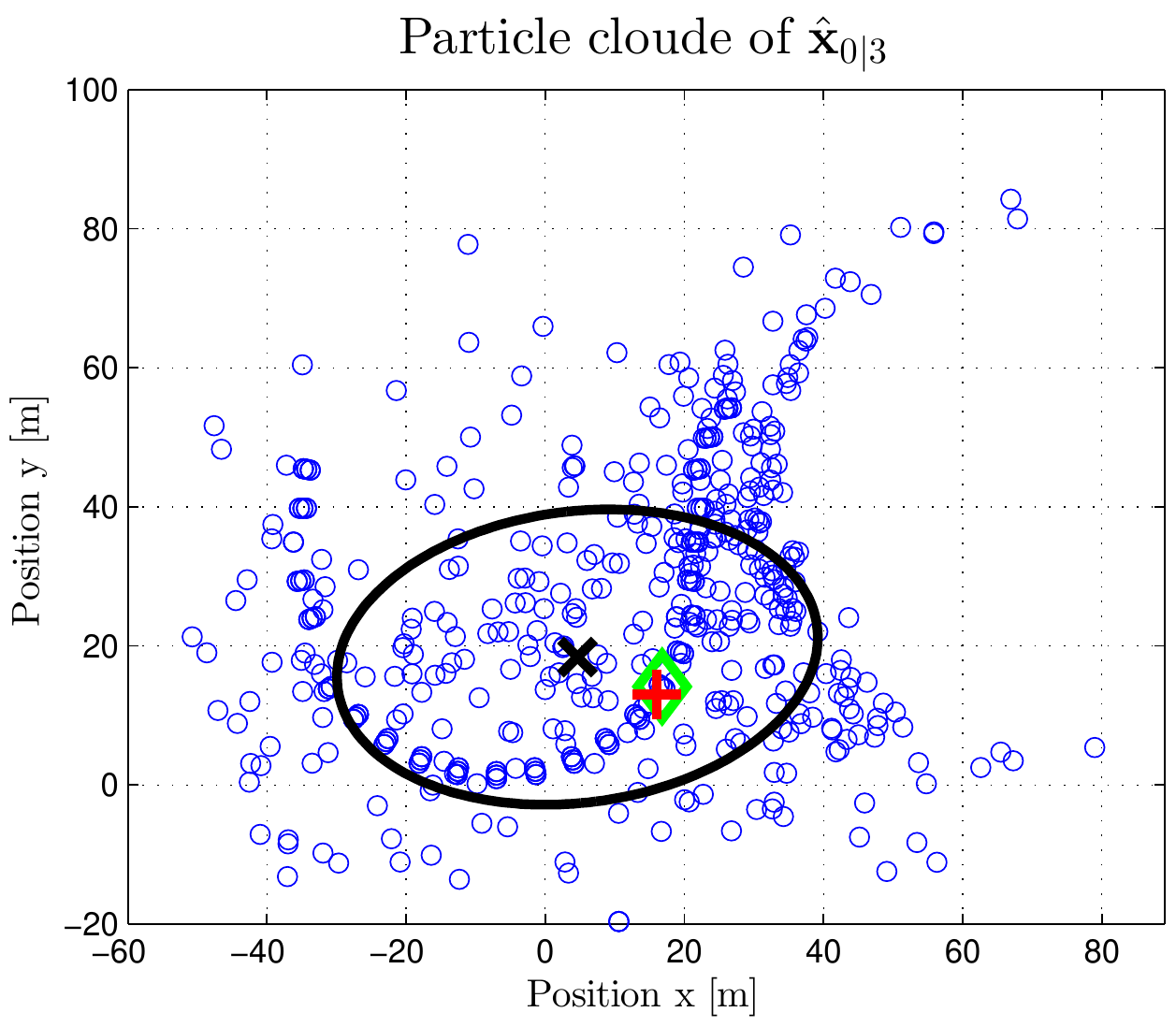}}

\hspace{1.5mm}
\resizebox{0.47\linewidth}{!}{\includegraphics{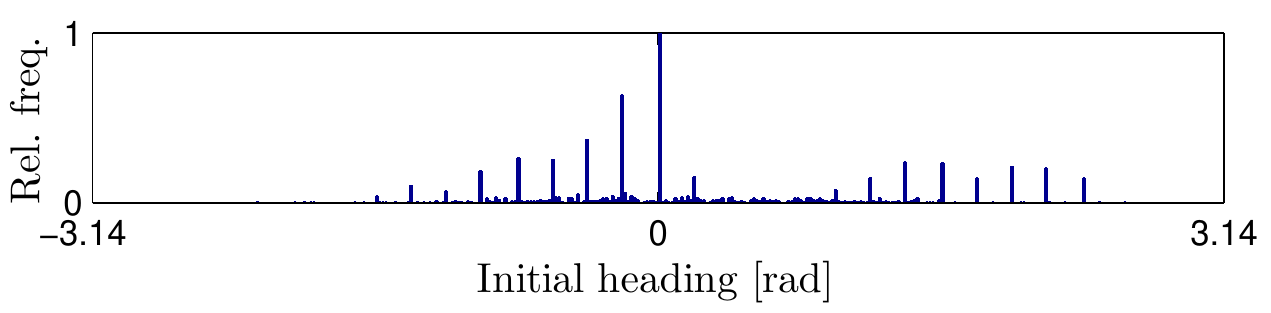}}
\resizebox{0.47\linewidth}{!}{\includegraphics{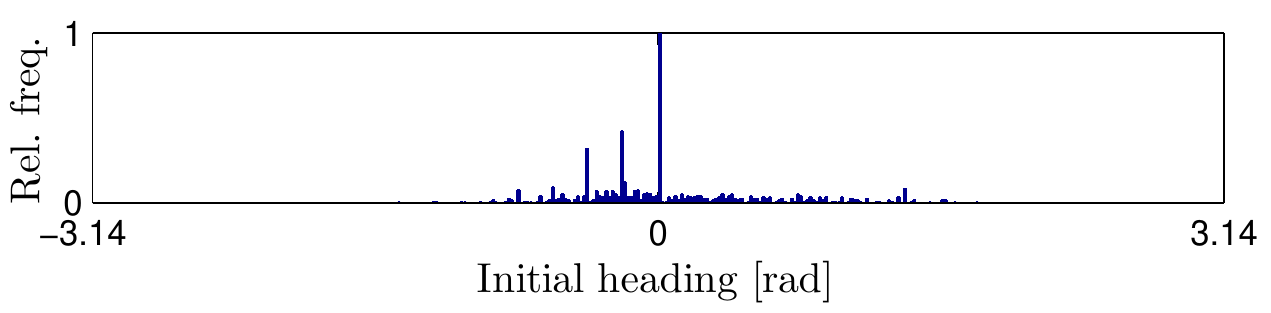}}

\vspace{4mm}
\resizebox{0.47\linewidth}{!}{\includegraphics{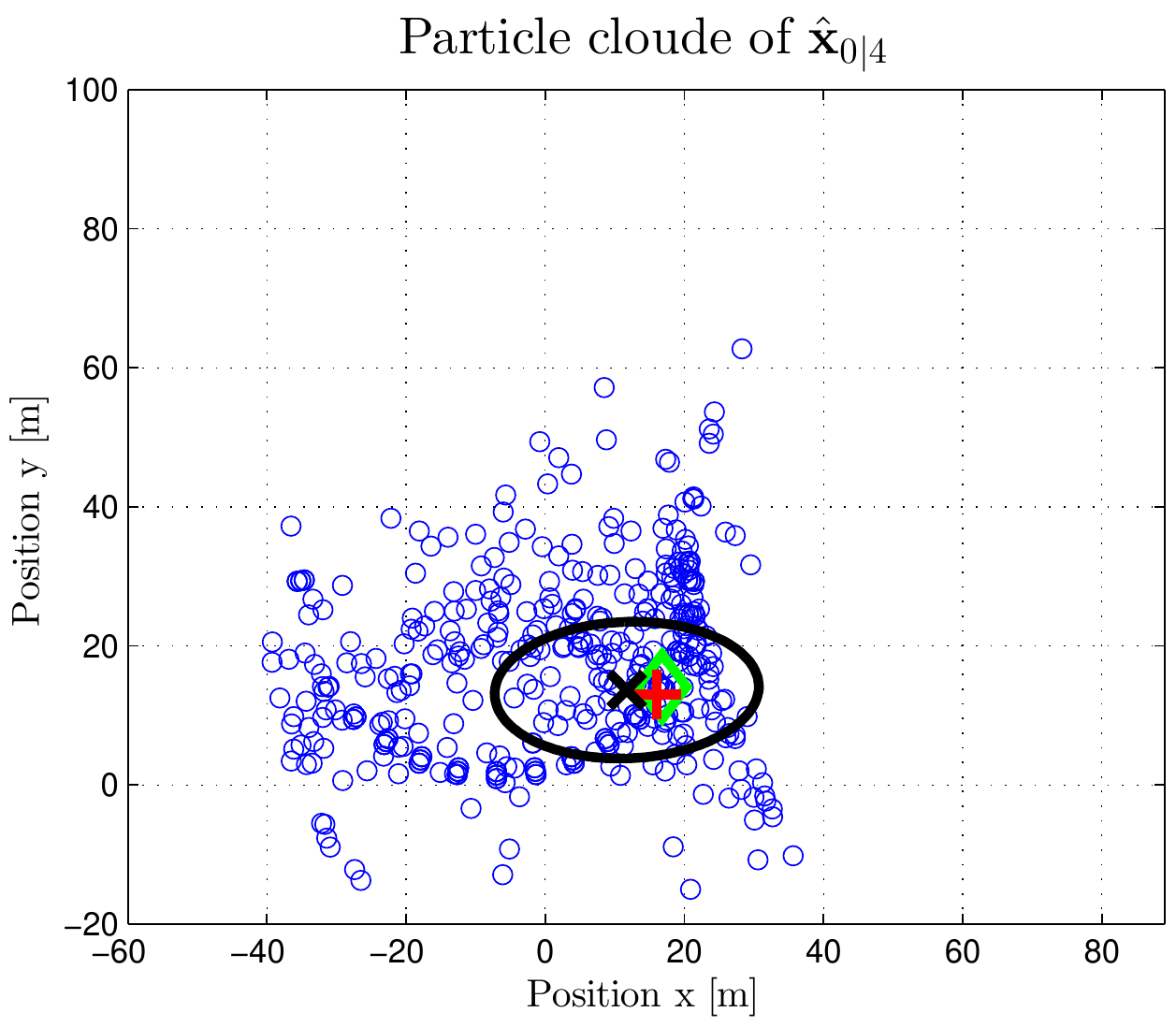}}
\resizebox{0.47\linewidth}{!}{\includegraphics{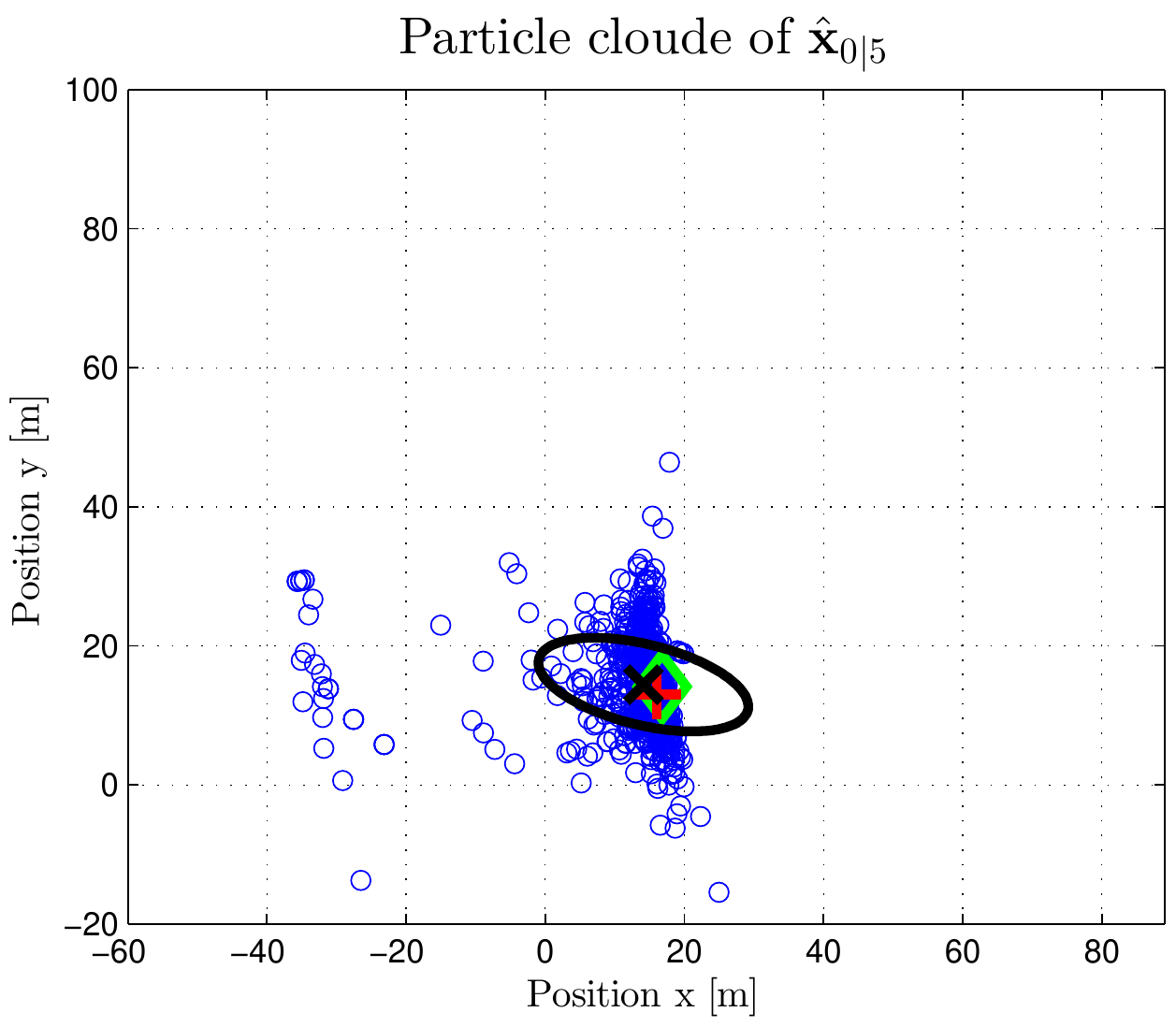}}

\hspace{1.5mm}
\resizebox{0.47\linewidth}{!}{\includegraphics{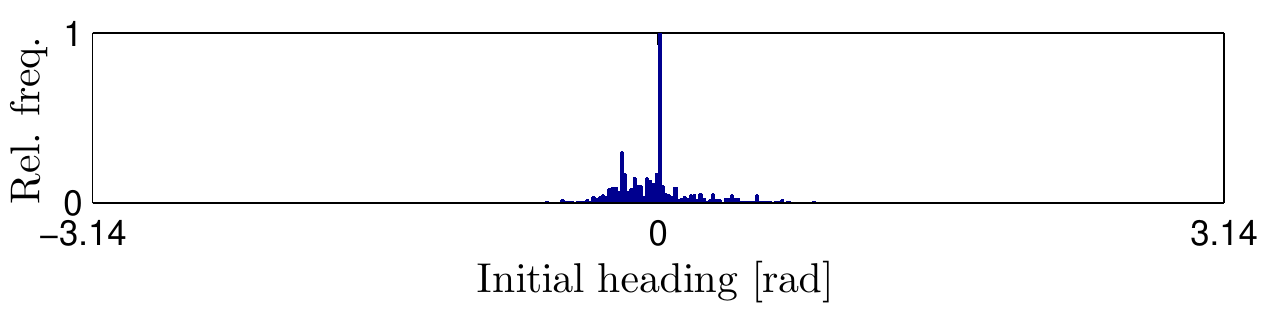}}
\resizebox{0.47\linewidth}{!}{\includegraphics{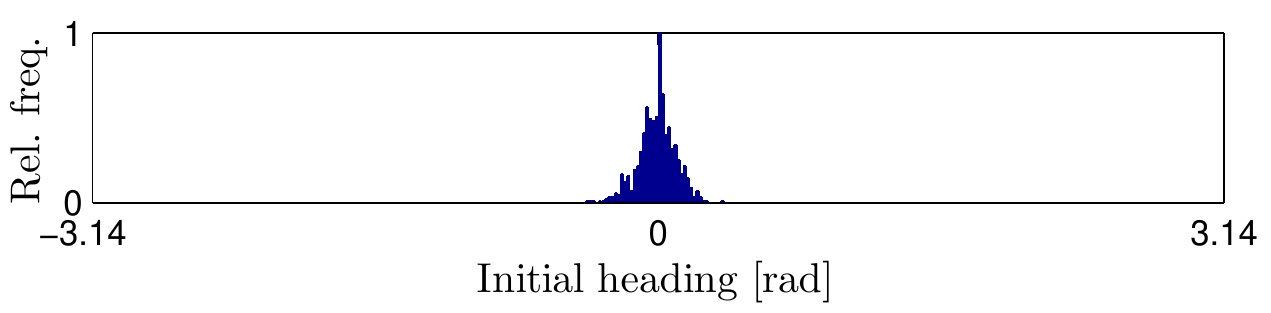}}

\vspace{4mm}
\resizebox{0.47\linewidth}{!}{\includegraphics{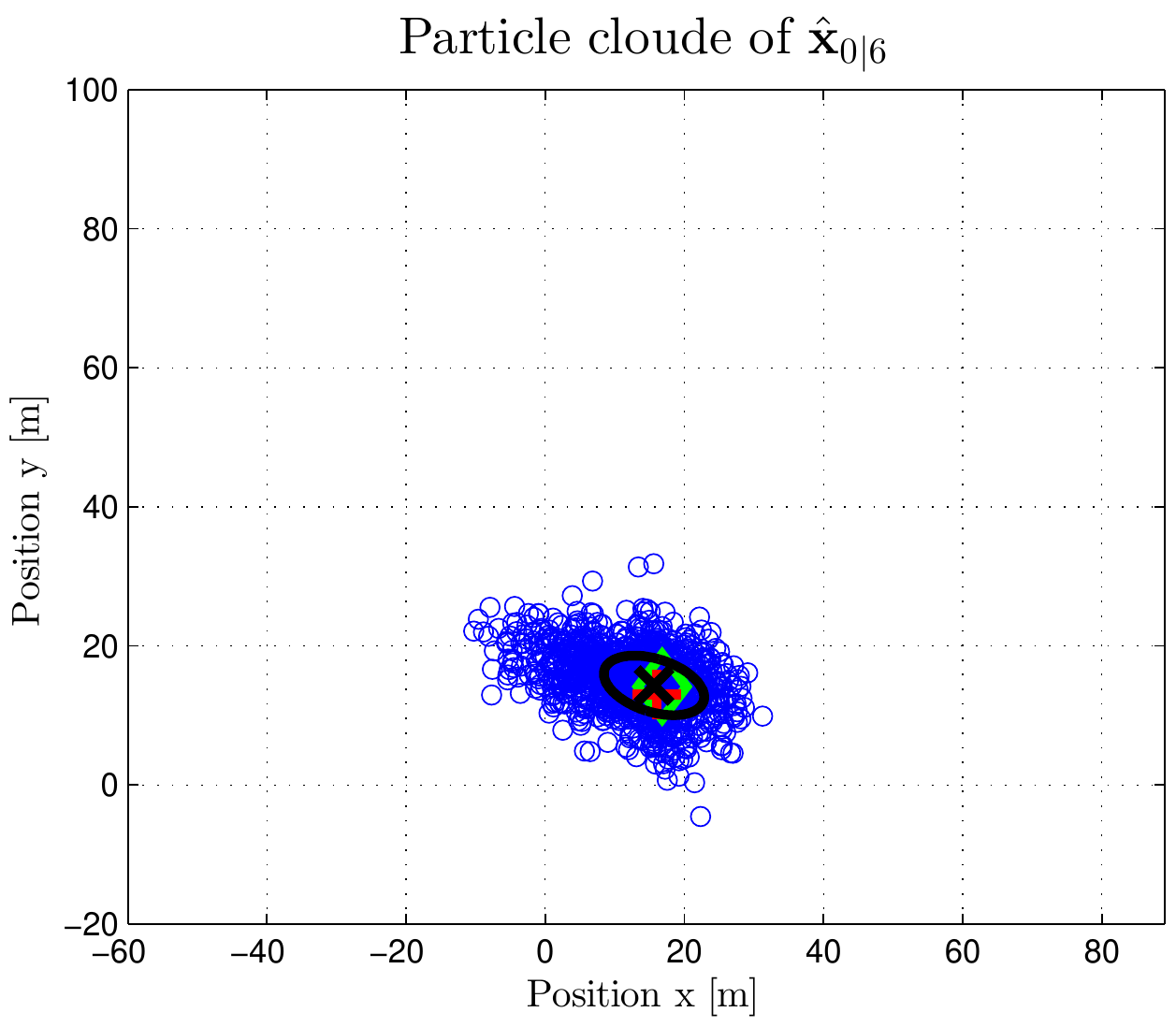}}
\resizebox{0.47\linewidth}{!}{\includegraphics{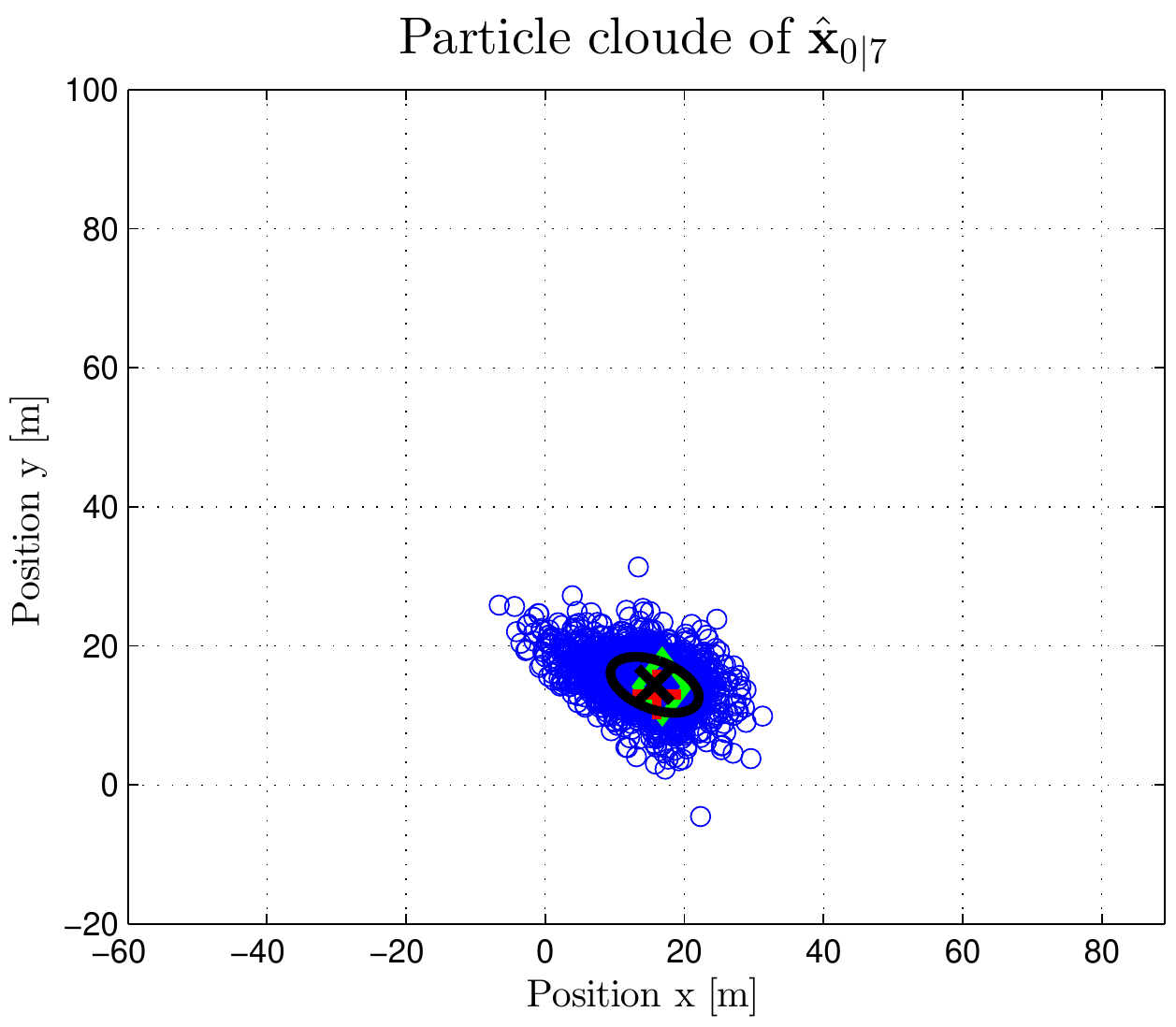}}

\hspace{1.5mm}
\resizebox{0.47\linewidth}{!}{\includegraphics{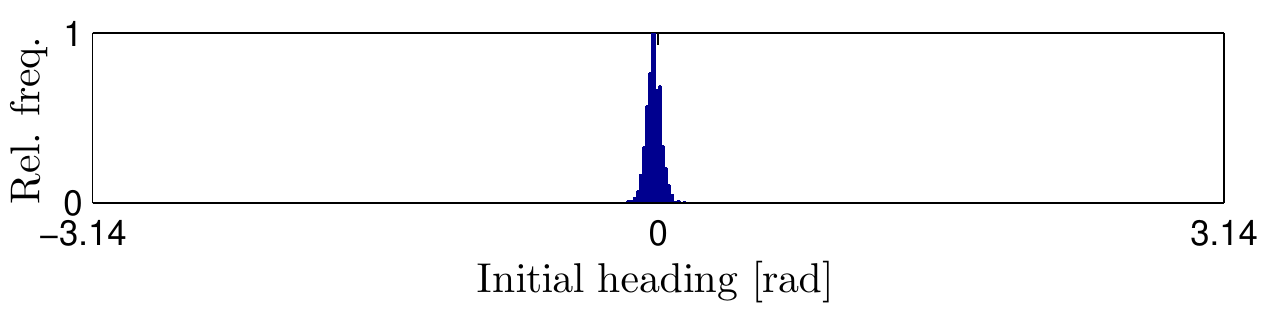}}
\resizebox{0.47\linewidth}{!}{\includegraphics{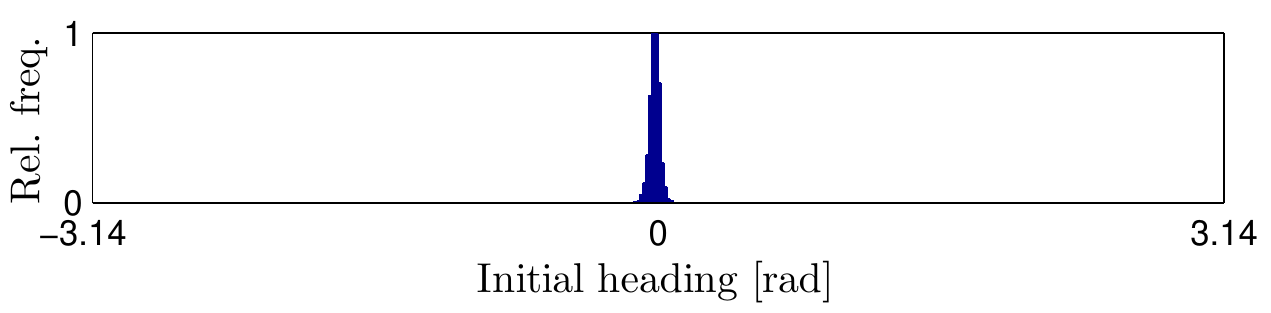}}
\caption{The plots show the particle clouds in the horizontal plane (particles with weights $w^{(i)}>1/N$) and the weighted histograms of particle headings after conditioning by the respective range measurements. The red plus-signs indicate the true initial position $\mathbf{x}_0$, the black crosses indicate the recursive estimates $\hat{\mathbf{x}}_{0|\ell}$, and the green diamonds indicate the most likely particle. The black ellipses indicate the one-sigma confidence ellipse given by $\mathbf{P}_{0|\ell}$.}
\label{fig:point_clouds}%
\end{figure}

To illustrate the Bayesian recursive initialization as described in the previous section, we simulate a basic scenario. Two agents move along the trajectories and perform the ranging shown in Fig.~\ref{fig:sim_trj}. The ranging errors are assumed Cauchy-distributed with scale $\sigma=1$~[m]. The lower agent is initialized relative the upper agent. A realization of the resulting recursive state estimates is seen in Fig.~\ref{fig:pos_est}. The related particle clouds (particles with weights $w^{(i)}>1/N$) in the horizontal plane and histograms of their headings are seen in Fig.~\ref{fig:point_clouds}. After the first ranging, the particles are evenly distributed in a torus. After the second ranging, only a few rings of particles remain. Following subsequent ranging, the particle clouds become more and more uni-modal and the mean $\hat{\mathbf{x}}_{0|\ell}$ and the most likely particle are seen to approach the true initial position. 
Finally, the covariance is low enough that the subsequent estimation can be performed by a covariance-based filter. 

\begin{figure}[t]%
\centering
\resizebox{\linewidth}{!}{\includegraphics{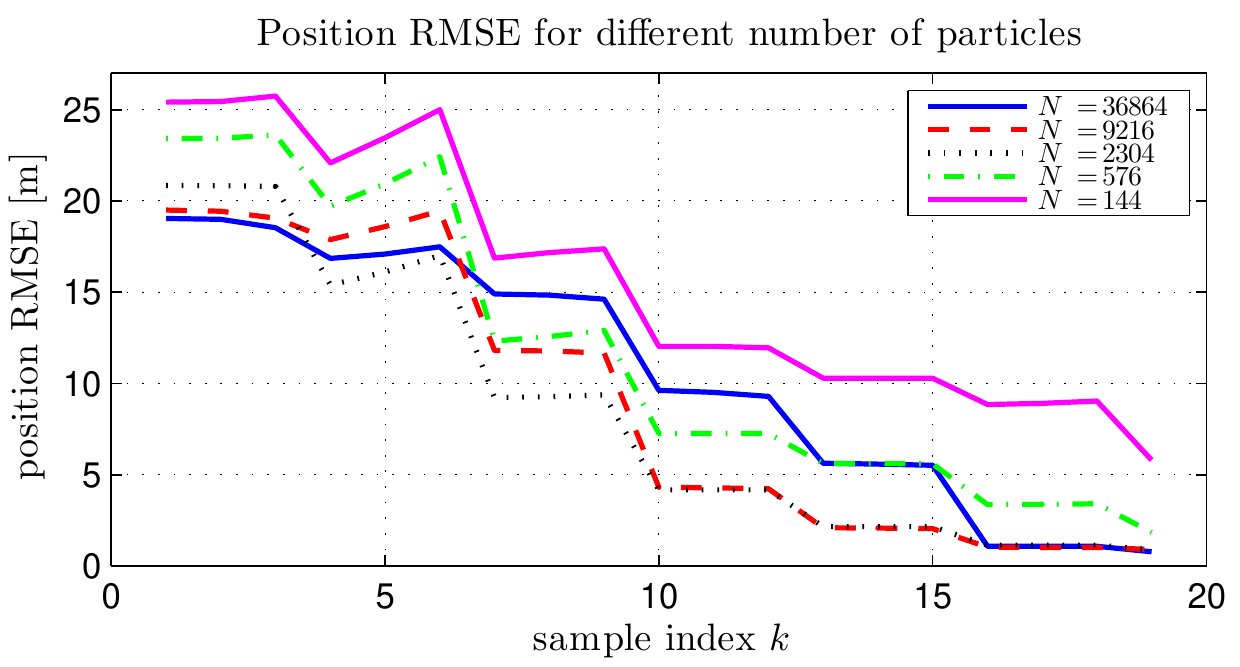}}
\caption{Position RMSE for 100 realizations during the initialization for different number of particles. As can be seen only a few hundred particles is required to get an acceptable performance for the simulated scenario.}\label{fig:pos_err_nr_hyp}
\vspace{-5mm}
\end{figure}

For clarity, in Fig.~\ref{fig:point_clouds}, an excessive number of particles are used ($N=9216$). However, the initialization results are largely unchanged for down to a few hundred particles. The position root-mean-square-error (RMSE) as a function of the sample index for different number of particles is shown in Fig.~\ref{fig:pos_err_nr_hyp}. The RMSEs have been calculated over 100 realizations with random seedings of the sampling. 
The number of particles is varied by changing the granularity of the base hypotheses of the heading and the bearing. The displayed numbers of particles $(36864,9216,2304,576,114)$ correspond to the granularities $(5.625^\circ,11.25^\circ,22.5^\circ,45^\circ,90^\circ)$. The initial range hypotheses are $(\tilde{r}_0-1,\tilde{r}_0,\tilde{r}_0+1)$ and the height hypotheses are $(-0.5,0,0.5)$. It is observed that the performance of the initialization is largely unchanged for granularities equal to or below $45^\circ$. 
The behavior displayed around $45^\circ$ and above should come as no surprise since clearly a granularity of $45^\circ-90^\circ$ is very coarse. However, note that the traveled distance is approx. 150~[m] so the final RMSE of 7~[m] for the 90$^\circ$ case is still acceptable.


\section{Experimental data}\label{sec:exp_data}
\begin{figure}[t]%
\centering
\begin{tikzpicture}[x=2.5mm,y=2.5mm]
\node {\includegraphics[clip=true,trim=15mm 0 11mm 0,angle=-90,width=0.85\linewidth]{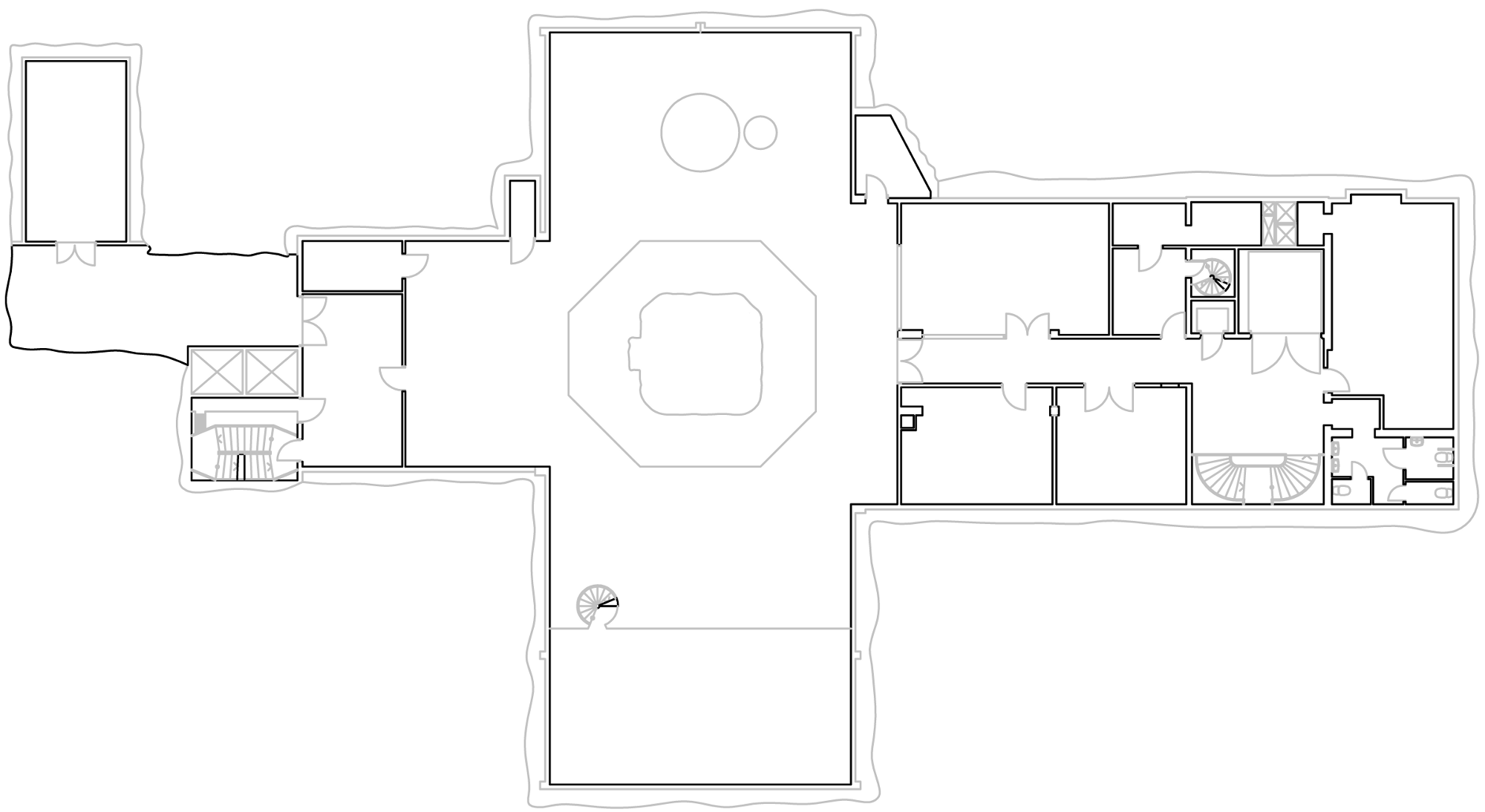}};
\draw[thick,->] (-42mm,-29mm) -- (40mm,-29mm) node[midway,below=0mm] {\footnotesize position $x$};
\draw[thick,->] (-39mm,-32mm) -- (-39mm,30mm);
\node at (-41.5mm,0) [rotate=90] {\footnotesize position $y$};
\node at (0,31mm) {\small Recursive initialization of two agents in R1};
\begin{scope}[yshift=-21mm,xshift=11mm,rotate=213,scale=0.93]
\draw[blue] plot[smooth] file {pos_agent1.txt};
\draw[red,dashed] plot[smooth] file {pos_agent2.txt};
\end{scope}
\end{tikzpicture}
\caption{Result of recursive initialization of one agent (red dashed line) relative to another agent (blue solid line) overlayed on a floor-plan. The estimated trajectories roughly correspond to the true trajectories.}\label{fig:exp_data}
\end{figure}
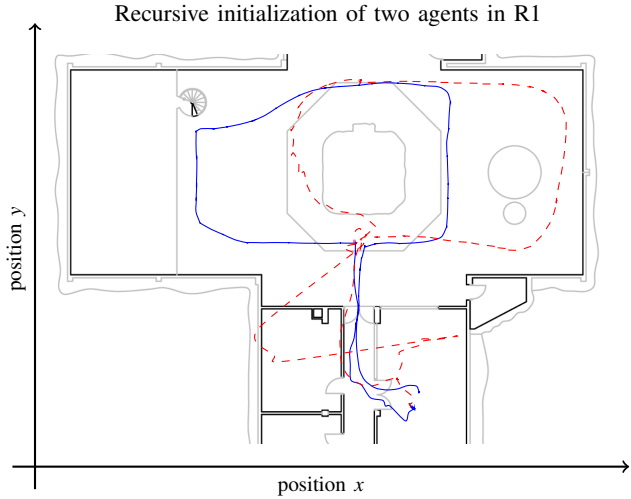

The recursive Bayesian initialization has been implemented as a part of the positioning system described in~\cite{Nilsson2013}. Fig.~\ref{fig:exp_data} shows the initialization (jagged trajectory segment) and subsequent covariance-based position estimation of one agent (red dashed line) relative to another agent (blue solid line) equipped with OpenShoe dead reckoning units~\cite{Nilsson2012} and with synthetic range measurements provided by the Ubisense real-time localization system installed in the R1 reactor hall. The trajectories roughly correspond to the true trajectories. Overall, the suggested initialization method has shown stable results over the development of the localization system.

\section{Conclusion}\label{sec:concl}
In this article, we have suggested a method for initializing the state estimation in a cooperative localization scenario based on dead reckoning and ranging. This is done by recursively estimating the initial state of an agent by a particle filter while treating its surrounding and its dead reckoning as deterministic. Estimating the initial state rather than the the current state has been shown to give an easier estimation problem requiring less particles and giving a lower computational cost. The effectiveness of the method has been demonstrated with simulations and a real system implementation.

\bibliographystyle{ieeetr}
\bibliography{init}

\end{document}